\documentclass[10pt,twocolumn,letterpaper]{article}

\usepackage{cvpr}
\usepackage{times,url}
\usepackage{epsfig}
\usepackage{graphicx}
\usepackage{amsmath}
\usepackage{amssymb,subfigure,tabularx,multirow,cite,enumerate}
\usepackage[table]{xcolor} 

\usepackage[breaklinks=true,bookmarks=false]{hyperref}
\cvprfinalcopy 

\def\cvprPaperID{****} 

\begin{document}

\title{Fast Color Constancy with Patch-wise  Bright Pixels}

\author{Yiyao Shi, Jian Wang\thanks{Correspondence to {\tt\small jian\_wang@fudan.edu.cn}} ~and Xiangyang Xue  \\ 
Fudan University \\
}

\maketitle

\begin{abstract}
		 In this paper,  a  learning-free color constancy algorithm called the Patch-wise Bright Pixels (PBP) is proposed. In this algorithm, an input image is first downsampled and then cut equally into a few patches. After that, according to the modified brightness of each patch,  a proper fraction of brightest pixels in the patch is selected. Finally,  Gray World (GW)-based methods are applied to the selected bright pixels to estimate the illuminant of the scene. Experiments on NUS $8$-Camera Dataset show that the PBP algorithm outperforms the state-of-the-art learning-free methods as well as a broad range of learning-based ones. In particular, PBP processes a $1080$p image within two milliseconds, which is  hundreds of times faster than the existing learning-free ones. Our algorithm offers a potential solution to the full-screen smart phones whose  screen-to-body ratio is $100$\%.   		 
\end{abstract}

\section{Introduction}

For decades, color constancy methods have been widely adopted to remove the color cast triggered by the light source, sensor sensitivity, surface reflection, etc.~\cite{GW,WP,SoG}. A natural way for color constancy is to  estimate the color of light source from the color-biased images, and then offset the color bias according to the estimation~\cite{comput}.  In general, the illuminant estimation methods can be grouped into two major categories: i) those relying on learning and (ii) those being learning-free. 
In a nutshell, the learning-based methods utilize the features of input images and the associated training data to learn the regressors for illuminant estimation. Representative algorithms include 
	 Bayesian Color Constancy~\cite{Bayesian}, 
	 Zakizadeh~{\it et~al.}~\cite{hybrid}, 
	 Gamut Mapping~\cite{Gamut}, 
	 Corrected Moment~\cite{Corrected Moment},
	 Convolutional Color Constancy (CCC)~\cite{CCC}, 
	 Fast Fourier Color Constancy (FFCC)~\cite{FFCC}, 
	 Chakrabarti~{\it et~al.}~\cite{Cha}, 
	 Fully Convolutional Color Constancy (FC$_4$)~\cite{FC4}, etc.  
	These methods generally have satisfactory performance, yet at the price of increased computational cost and specific training for each camera or dataset. In FFCC~\cite{FFCC}, for instance, an image is transformed into a $\log$-chroma histogram, filtered by a convolutional kernel learned from each dataset, and finally fitted in a bivariate von Mises distribution~\cite{Distribution}. While FFCC has been one of the best methods in the learning family, its performance is fundamentally limited when the training and test sets are from disparate datasets~\cite{Gray Index}.


%
%
%
%
%
%
%
	
	\begin{figure}[t]
\centering
{\includegraphics[width = \linewidth] {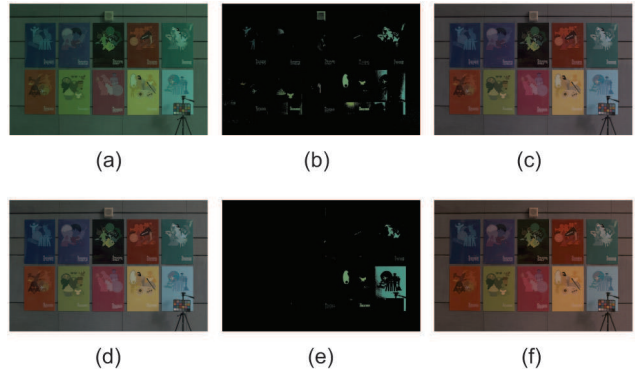} \label{pixel and patchf}}	
		\caption{An example for the performance comparison of PBP and BP~\cite{Bright Pixels}: (a) input image; (b) selected pixels for PBP; (c) image corrected with PBP (angular error $=3.13^\circ$); (d) image corrected with groundtruth illuminant; (e) selected pixels for BP; (f) image corrected with BP (angular error $ = 8.24^\circ$). A gamma correction with $\gamma = 2.2$ is used to enhance the overall brightness of images.}
		\label{pixel and patch}
\end{figure}

	In the second category are learning-free methods. In contrast to those learning-based ones, the learning-free  methods directly predict the illuminant from information  collected  from the input image.  Since they work independent of  both the camera information and training data, their implementation is  much  easier and faster.  Algorithms in this category contain 
 Gray World (GW)~\cite{GW},   White Patch (WP)~\cite{WP},  Shades of Gray (SoG)~\cite{SoG},   Gray Edge (GE)~\cite{GE}, Local Surface Reflectance Statistics (LSRS)~\cite{LSRS}, Cheng~{\it et~al}.~\cite{PCA},  Bright Pixels (BP)~\cite{Bright Pixels},  Gray Pixels (GP)~\cite{Gray Pixels},  Gray Index (GI)~\cite{Gray Index}, etc.   
By assuming the corresponding reflectance in a scene to be achromatic, the GW-based methods (e.g.,~\cite{GW, WP, SoG, GGW, GGW2, GE}) compute the $p$th Minkowsky-norm of each color channel in the input image  (or the derivative)   as the estimated illuminant. For example, WP takes the maximum response ($\ell_{\infty}$-norm) of the RGB-channel as the estimated illuminant under the assumption that the maximum reflectance in a scene is achromatic~\cite{WP}.

	As a natural extension to the WP method, BP~\cite{Bright Pixels} selects the brightest $\sigma\%$ pixels (henceforth called the bright pixels), and then applies the GW-based methods on the selected pixels to estimate the illuminant.  The bright pixels are helpful for illuminant estimation  because the bright areas of images are usually light sources, highlights, specularities and white surfaces, which are highly effective in reflecting the information of light source. Indeed, in~\cite{Bright Pixels} the performance gain was clearly observed when the GW-based methods were applied to the bright pixels, rather than  over all pixels. 
However, BP focuses only on the top-brightness pixels of images while neglecting the spatial information of scenes. For input images with multiple light sources or large non-white bright areas, therefore, BP may not exhibit promising performance; See Figures~1(e) and 1(f) for example.

	In this paper, we propose a fast color constancy algorithm called the Patch-wise Bright Pixels (PBP), which identifies  bright pixels in a patch-wise manner. Specifically, we first divide each input image into tens or hundreds of square patches and collect their center pixels to form the downsampled image. Then, by equally dividing the downsampled image into multiple patches, we select a proper fraction of brightest pixels in each patch. Finally, we estimate the illuminant based on the totally selected bright pixels. In this way, the selected bright pixels are well scattered throughout the entire image plane, thereby reducing the potential estimation failures due to too concentrated selection of local bright areas; See Figures~1(b) and~1(c).  
	
The main contribution of our paper is twofold. 

\begin{enumerate}[i)]
\item For the sake of computational efficiency, we propose a downsampling pretreatment on the input image. While a massive reduction on the running time is achieved, the downsampling operation nevertheless preserves  the accuracy of  illuminant estimation for a broad range of learning-free methods.  

\item Through experiments over the NUS $8$-Camera dataset~\cite{PCA}, it is demonstrated that the proposed PBP method has superior performance compared to the state-of-the-art learning-free methods (e.g., GI~\cite{Gray Index}), yet is hundreds of times faster. Therefore, our PBP algorithm allows the real-time processing of the full high definition (HD) videos, which offers a potential solution to the design of full-screen smart phones whose  screen-to-body ratio is $100$\%.   
\end{enumerate}

\section{Related Work}

In the learning-free category, 	GW~\cite{GW} and WP~\cite{WP} take the mean and maximum response (i.e., the Minkowski $\ell_1$- and $\ell_{\infty}$-norm) of the RGB-channel as the estimated illuminant, respectively. Methods based on other norms have also been discussed; See, e.g.,~\cite{SoG}. Among those, the $\ell_6$-norm based method generally achieves the best performance~\cite{SoG}. 
On top of that, the GW-based methods incorporating more sophisticated operations have also been studied. For example, prior to calculating the Minkowski norms, the general GW (GGW)~\cite{GGW, GGW2} employs a local smoothing operation; Whereas, GE~\cite{GE} exploits extra information by computing the first- or second-order derivative. Both GP~\cite {Gray Pixels} and GI~\cite{Gray Index} apply the conventional GW methods on the selected gray pixels (i.e., those with roughly the same response in the RGB-channel). In addition, other methods performing the illuminant estimation task based on  the local surface reflectance statistics have  been suggested; See LSRS~\cite{LSRS}.

	Since the WP~\cite{WP} method focuses only on the maximum response of color channels, it may not be robust to noise. To improve the robustness,~\cite{WP1, WP2} execute
	 a local smoothing operation on the input images. BP~\cite{Bright Pixels} extends the WP method by selecting a fraction of brightest pixels.  The WP Gamut algorithm, introduced by Joze~{\it et~al.}~\cite{WP Gamut}, applies the Gamut mapping to the top-brightness pixels only, which outperforms the global-wise Gamut Mapping methods.  Furthermore, by selecting both the brightest pixels as well as the darkest ones, Cheng {\it et~al.}~\cite{PCA} takes the first eigenvector of the covariance matrix corresponding to the RGB-channel of selected pixels as the estimated illuminant. So far, Cheng {\it et~al.}~\cite{PCA} has been one of the best-performing color constancy algorithms in the learning-free family. 


\section{Downsampling} \label{sec:downsampling}
\subsection{Overview}

Over the years, downsampling has been widely used in electrical engineering and computer science. In many learning-based color constancy methods~\cite{Shi2016, FFCC, Cha, Bianco1, Bianco2, FC4, Recurrent}, it has also been used to fit the input images into model, prevent overfitting, improve generalization, reduce the computational cost, etc. For instance, in FFCC-full~\cite{FFCC}, the input images are normalized to $384 \times 256$ ones so as to fit the input size of algorithm. Also, FFCC-thumb downsamples each input image to a thumbnail patch with size $48\times 32$ to expedite the learning process. Despite these advantages, the learning-based approaches with various sizes of input patches have to be trained individually, whose performance could be massively impaired when the sampling rate is low.

To date, little has been known about the application of downsampling to the learning-free approaches. Nevertheless, downsampling can be of great help to these methods. Indeed,  since the computational cost of the learning-free algorithms mostly scales linearly in the image size,  the application of downsampling to them can significantly improve their computational efficiency. It is interesting, however, to note that the performance of learning-free algorithms is rarely  weakened by the downsampling operation. 
The reason is that a large proportion of learning-free algorithms focus merely on the overall color information. While downsampling causes the loss of spatial and color information of images, the missing information does not matter much to the performance of illuminant estimation.

\subsection{Metric and Dataset}
To study the practical effect of downsampling, we use an experiment to examine the performance of  learning-free methods under various downsampling rates. We follow the strategy of~\cite{Angular Error}, in which the effectiveness of illuminant estimation algorithms is evaluated by checking the angular error between the  estimated illuminant  direction $\mathbf{e}_{est}$ and the groundtruth illuminant direction $\mathbf{e}_{gr}$ of an input image:
	\begin{equation}
		angular~error:= \cos^{-1}\left(\frac{\mathbf{e}_{est} \cdot \mathbf{e}_{gt}}{\|\mathbf{e}_{est}\|_2\|\mathbf{e}_{gt}\|_2}\right). 
	\end{equation}
For a set of input images, we are often interested in the mean and median angular errors. These metrics allow to empirically compare the performance of different  approaches. 

 In our experiment, we consider two popular datasets: 
    \begin{enumerate}[i)]
		\item NUS $8$-Camera Dataset~\cite{PCA}:  $1736$ high dynamic linear images taken from eight cameras;
		\item Gehler-Shi Dataset~\cite{Gehlershi}:   $568$ high dynamic linear images taken from two cameras.
	\end{enumerate}
Some pre-processing operations are needed for images in these datasets. Specifically, to properly estimate the illuminant of an image, we have to carefully clip the pixels whose light reflectance exceeds the dynamic range of the camera. For images in the Gehler-Shi Dataset~\cite{Gehlershi}, we remove those pixels that exceed $95$\% of the maximum response for any color channel. Whereas for the NUS $8$-Camera Dataset~\cite{PCA}, the threshold is set to be $97$\%. Besides, we uniformly transform the images in both datasets from $12$-bit (or $14$-bit) to $8$-bit. This is mainly because the input images of  $8$-bit roughly lead to the same performance as the original ones, yet with much shorter running time.

\subsection{Experimental Result}
		\label{sec:exp3.3}
		
%
%
%
	

In our experiment, we apply equidistant downsampling with various intervals to both rows and columns of each input image.  For instance, if the downsampling interval is five, then the image is divided into hundreds of  patches of size $5 \times 5$. In each patch, only the pixel located at the central point is sampled. 	For comparitive purpose, the following representative approaches are included in our experiment: WP~\cite{WP}, GW~\cite{GW}, SoG~\cite{SoG}, GGW~\cite{GGW}, First-order Gray-Edge (GE$_1$)~\cite{GE}, Second-order Gray-Edge (GE$_2$)~\cite{GE}, LSRS~\cite{LSRS}, Cheng {\it et~al}.~\cite{PCA} and GI~\cite{Gray Index}. Each approach is performed on the downsampled images with downsampling interval ranging from $1$ to $20$. The parameters of these approaches are specified in Table~\ref{downsample_parameter}, if any.

	\begin{table}[t]
		\centering
		\caption{Parameters of Testing Algorithms}
		\begin{tabular}{c|c}
		\hline
			{\bf Algorithm} & {\bf Parameters}
			\\ \hline   
			SoG~\cite{SoG} & $k=0$, $p=7$, $\sigma=0$ \\\hline
			GGW~\cite{GGW} & $k=0$, $p=11$, $\sigma=1$ \\\hline
			GE$_1$~\cite{GE} & $k=1$, $p=7$, $\sigma=1$ \\\hline
			GE$_2$~\cite{GE} & $k=2$, $p=7$, $\sigma=1$ \\\hline
			LSRS~\cite{LSRS} & $3 \times 4$  patches \\\hline
			Cheng~{\it et~al.}~\cite{PCA} &$ n = 3.5\%$ \\\hline
			GI~\cite{Gray Index} & $N=0.1\%$, $\epsilon = 10^{-4}$ \\\hline
		\end{tabular}
	\label{downsample_parameter}
	\end{table}

	\begin{figure*}[t]
		\centering
		\includegraphics[width=\linewidth]{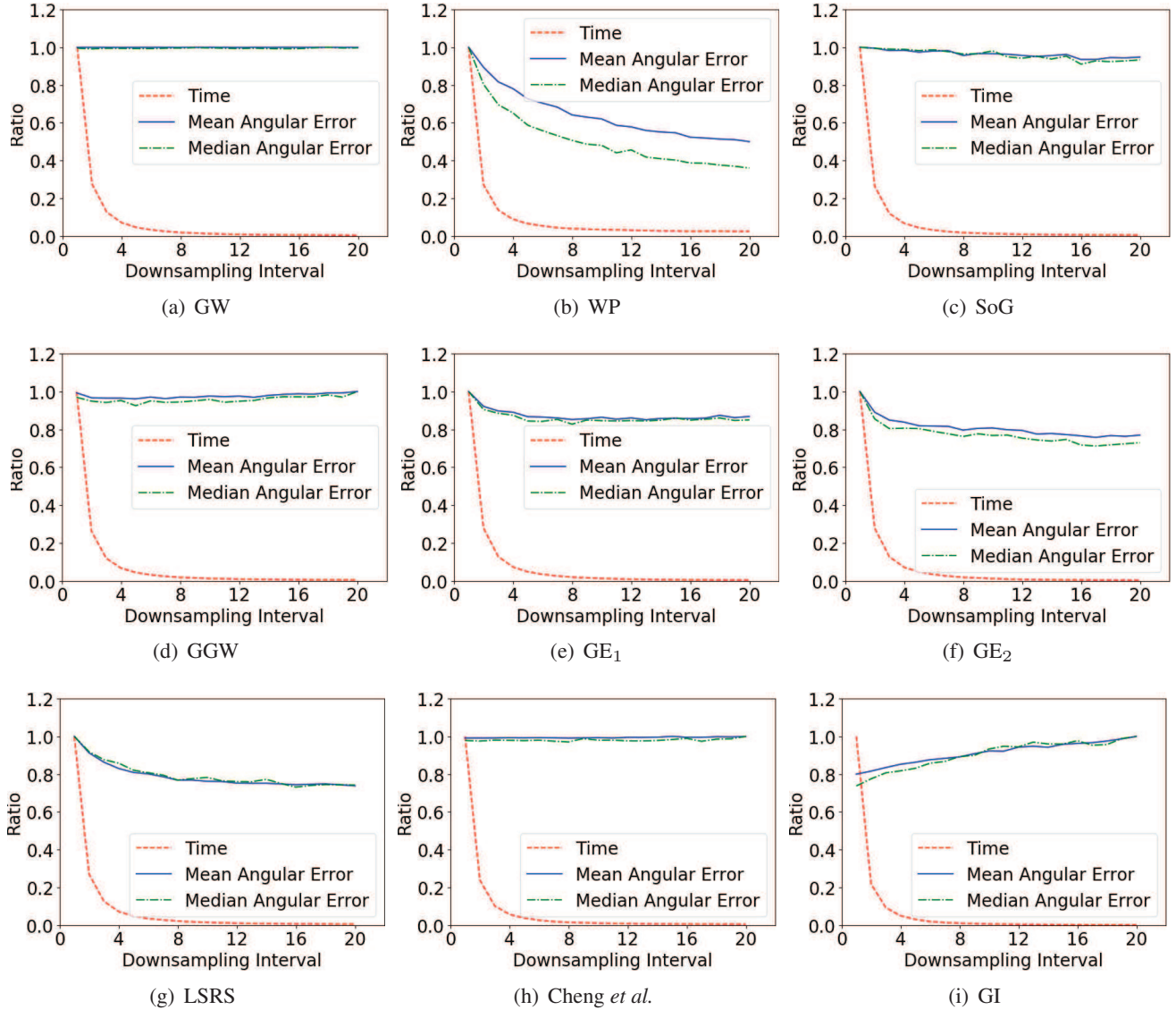} 
		\caption{The ratio of running time,  mean and median angular errors as a function of downsampling interval.}
		\label{downsample}  
	\end{figure*}

	The experiment is carried out on the NUS $8$-Camera Dataset~\cite{PCA}. The running time, mean and median angular errors as a function of the downsampling interval is displayed in Figure~\ref{downsample}. The result for each metric is normalized by setting the maximum value to one. In doing so, the vertical axis only represents the ratio whose unit is offset, yet the variation trend can be highlighted. It can be observed that downsampling is very effective in improving the computational efficiency of algorithms under test. Roughly speaking, for each testing algorithm, the running time scales inversely with the square of the downsampling interval,  which well matches our expectation,  given that the testing algorithms have a linear time complexity with respect to the number of pixels in the input image.

   In Figure~\ref{downsample}, one can notice the trend of mean and median angular errors for different approaches as the downsampling interval increases. Overall, for most of the testing methods, the mean and median angular errors do not increase with the downsampling interval. In particular, there is substantial decrease in both angular errors for the WP method as the downsampling interval goes large. This is because WP takes the maximum response of the RGB-channel to predict the illuminant, which, however, could come from the saturated pixels. Whereas if performed on the downsampled images, it is highly likely that WP takes a much lower response for each color channel, thereby reducing the risk of selecting the saturated pixels. The same explanation also applies to the performance enhancement of LSRS~\cite{LSRS}, as it takes advantage of the local maximal responses. 

The performance improvement of the  GE$_1$ and GE$_2$~\cite{GE} methods in the downsampled case  may be attributed to that the image derivative, based on which the illuminant is predicted, is enlarged by downsampling. This finding bears a resemblance to that in Cheng {\it et~al.}~\cite{PCA}, which manifests that the implementation of large artificial  gradients improves the performance of the GE method. For those using global color information (e.g., GW, SoG, GGW and Cheng {\it et~al.}), their performance generally remains unchanged whether the downsampling interval varies or not. On the other hand, it can be observed that the downsampling operation leads to a slight performance degration for GI~\cite{Gray Index}, which probably boils down to the following two reasons. Firstly, GI selects only a small fraction of pixels ($0.1$\%) in an image for illuminant estimation, which makes it vulnerable to the resolution reduction of the input image. Secondly, GI utilizes a mean filtering for the gray index map, thus further enhancing its susceptibility to the increased downsampling interval. In summary, 	the above experiment serves to underscore that the equidistant downsampling is effective in reducing the running time, while maintaining or even improving the performance of most learning-free methods  for illuminant estimation.

%
%

\section{ Patch-wise Bright Pixels}
%
	In this section, we first show the validity of bright pixels in reflecting the light source information and then put forward the PBP algorithm.
	
	\subsection{Why Bright Pixels?}
	The brightness $\bar{L}$ of a pixel is usually defined as the sum of its RGB-channel responses: 
	\begin{equation} 
	\bar{L} = R+G+B.
	\end{equation}
	In~\cite{Bright Pixels}, it has been revealed that bright pixels are helpful for illumiatant estimation. In the Gehler-Shi Dataset~\cite{Gehlershi}, for example, the groundtruth illuminant falls inside the gamut of the brightest $5$\% of pixels for more than $70$\% of images.  For this dataset, it has also been shown that there is a reduction of up to $50$\% on the median angular error when the GW-based methods (GW~\cite{GW}, GE~\cite{GE} and SoG~\cite{SoG}) are performed over the top $20$\% brightness pixels, rather than over all pixels of the input image  (see~\cite{Bright Pixels} for more details).

	To further investigate the effect of pixel brightness on the performance of illuminant estimation, we design an experiment concerning the angular error for pixels with different brightness. First, we sort the pixels of an image in ascending order of their brightness. Next, we evenly divide the sorted pixels into $100$ groups, where the $n$th group of pixels consists of the darkest $n$\% to $(n+1)\%$ ones. Then, for each group, we take the per-channel mean of pixels as the estimated illuminant:
	\begin{equation}
		I_c = \sum_{k\in P_i} \frac{I_{k,c}}{|P_i|}, \quad c\in\{R,G,B\},
	\end{equation}
	where $P_i$ denotes the set of pixels in the $i$th group, $|P_i|$ represents the cardinality of set $P_i$, and $I_{k,c}$ is the response of color channel $c$ for pixel $k$. Finally, the angular error is calculated based on the groundtruth illuminant and the estimated one. The experiment is carried out both on the NUS $8$-Camera Dataset~\cite{PCA} and  Gehler-Shi Dataset~\cite{Gehlershi}.
	
	\begin{figure}[t]
		\centering
		\subfigure[NUS $8$-Camera]{
			\begin{minipage}[t]{0.45\linewidth}
				\centering
				\hspace{-3mm}\includegraphics[width=\linewidth]{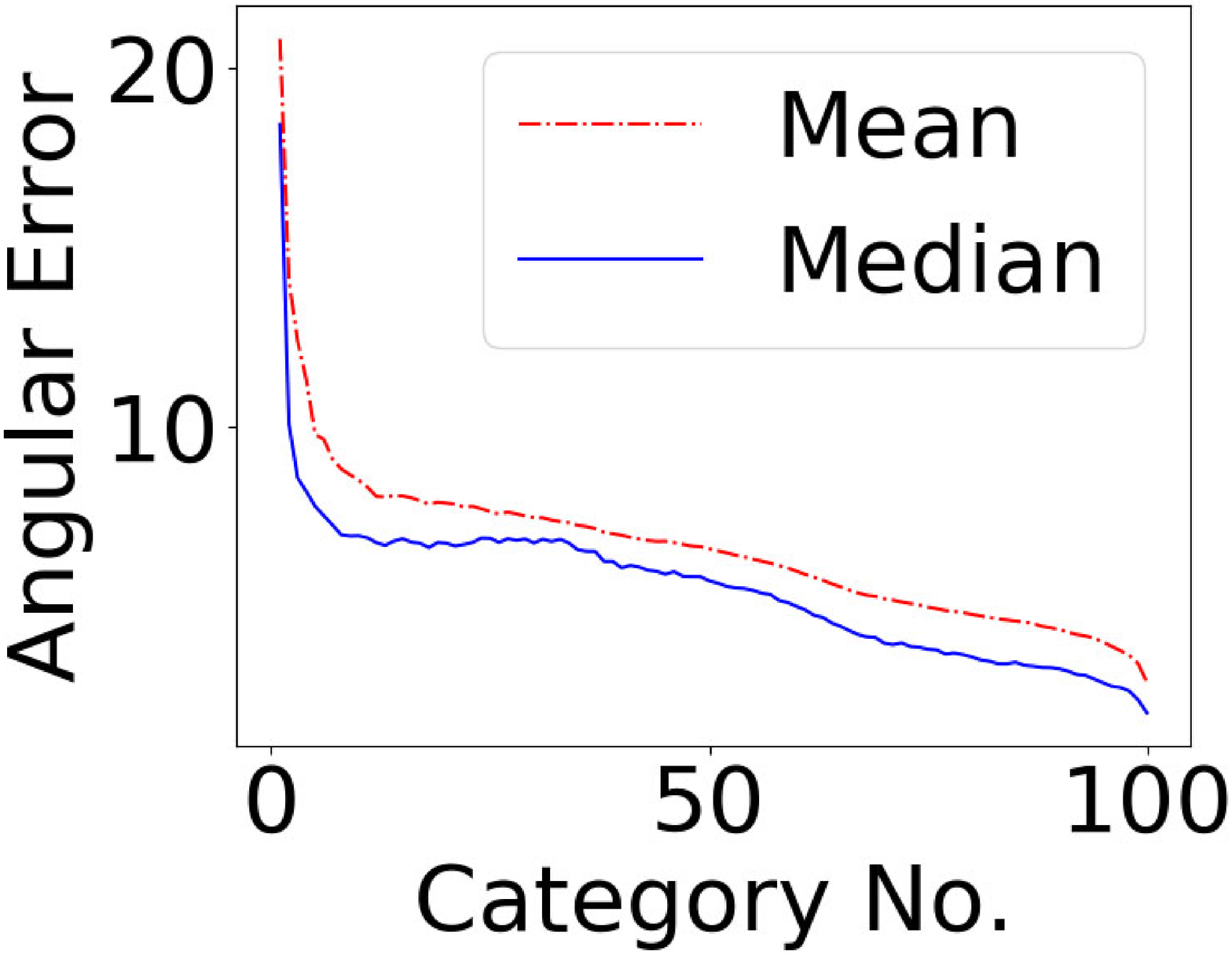}
			\end{minipage}}
		\subfigure[Gehler-Shi]{
			\begin{minipage}[t]{0.45\linewidth}
				\centering
				\includegraphics[width=\linewidth]{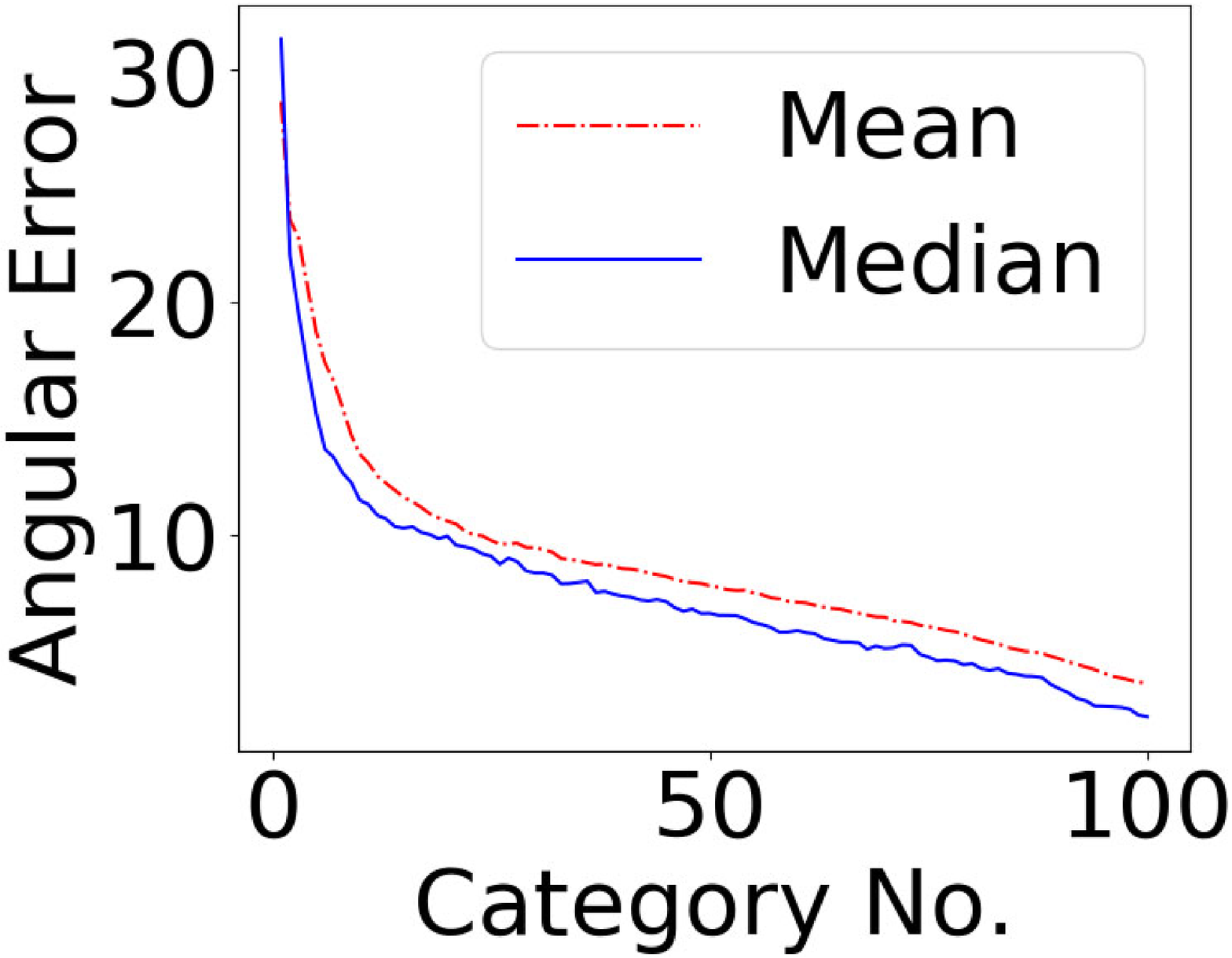}
			\end{minipage}} 
		\caption{Mean and median angular errors for $100$ groups of pixels.}
		\label{100_part}  
	\end{figure}

\begin{figure}[t]
\centering
\includegraphics[width = \linewidth]{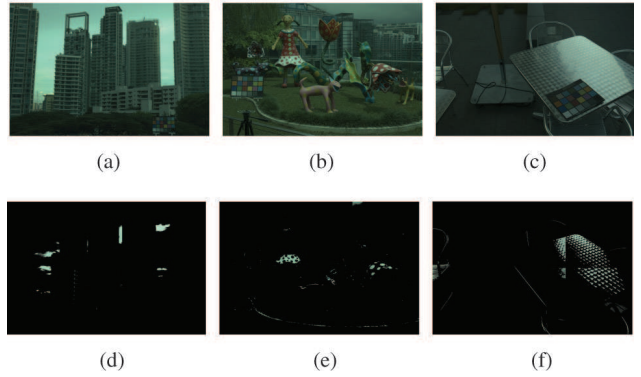} \label{4.6} 
		\caption{Examples in the NUS $8$-Camera Dataset~\cite{PCA}: (d) light sources, (e) white surfaces and (f) specularities are the selected bright pixels (with sampling rate  $3$\%) of the original images (a), (b) and (c), respectively. A gamma correction with $\gamma=2.2$ is applied to enhance the overall brightness of these images.  }
		\label{BP examples} 
\end{figure}

	The results of this experiment are shown in Figure \ref{100_part}.  For both datasets, a significant downward trend can be observed for the mean and median angular errors as the overall brightness of the sampled pixels increases. Accordingly, the outcome of our experiment, as well as the findings in~\cite{Bright Pixels}, both suggest that the bright pixels have great advantage over dark ones in predicting the illuminant of a scene. This is because the bright areas often incorporate highlights, specularities, white surfaces and light sources (see Figure \ref{BP examples} for example), which have remarkable effectiveness in reflecting the information of light source. Hence, utilization of bright pixels can be particularly helpful for illuminant estimation.

	\begin{figure*}[t]
		\centering
		\includegraphics[width=\linewidth]{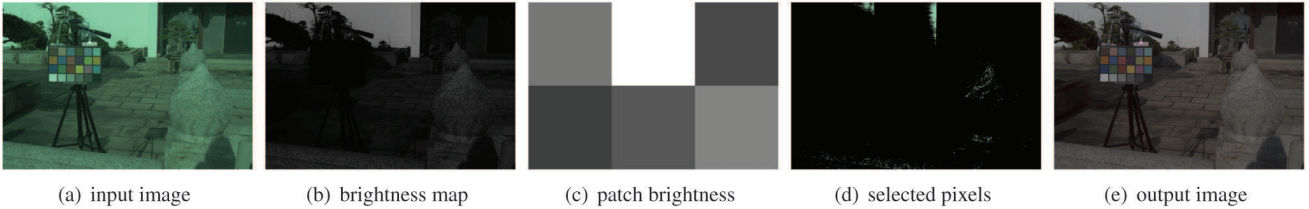}
	
		\caption{An example for the main steps of PBP. We set the downsampling rate to one for display purpose only; otherwise, the downsampled image could be too small to display here. From the input image (a), a brightness map (b) is first obtained by calculating $L/3 = (R+G+B)/3$ for each pixel. The image is then cut into $2\times 3$ patches with equal size. (c) is the sketch map for the modified brightness of each patch, which is the sum of pixel brightness in the patch (i.e., $q = 1$). After that, we set $\sigma =1\%$ and select a number of pixels (d) for each patch according to the modified brightness of the patch. Finally, we take the per-channel mean of selected pixels as the estimated illuminant. The output image (e) is the quotient of input image and the estimated illuminant. A gamma correction with $\gamma=2.2$ is used to (a), (d), (e) to enhance the overall brightness. }
		\label{illustration}
	\end{figure*}

	\subsection{The PBP Algorithm} \label{sec:algbpb}
	
\paragraph{Overview}
	In this section, we introduce the PBP algorithm, which performs color constancy based on both the bright pixels and patch brightness. In general, our algorithm consists of three major steps: i) downsample an input image, ii) select representative pixels from the downsampled image and iii) apply the GW-based methods to the selected pixels for illuminant estimation. 
	
\paragraph{Downsampling Input Image}	
	Thus far, the findings in Section~\ref{sec:downsampling} have demonstrated that downsampling can effectively reduce the running time of a majority of learning-free methods without even marginal performance degradation. This motivates to employ downsampling as a cost effective pretreatment for our algorithm.
Specifically, following the strategy in Section~\ref{sec:exp3.3}, we apply equidistant downsampling  with downsampling interval $S$ to the input image.

\paragraph{Selecting Representative Pixels}	
	Similar to GP~\cite{Gray Pixels}, GI~\cite{Gray Index} and BP~\cite{Bright Pixels}, the core of the PBP algorithm is the selection of representative pixels that are helpful for illuminant estimation. The main steps are as follows.
\begin{enumerate}[i)]
\item {\color{black}{We first}} set a rate $\sigma \in (0, 1)$. The number $N_{\sigma}$ of selected pixels can be given by:
	\begin{equation} \label{eq:sigma}
		N_{\sigma} = N \sigma,
	\end{equation}
	where $N$ is the total number of pixels in image $F$. The {\it modified brightness} $L$ of the image is computed as the sum of the $q$th power of all pixel brightness:
	\begin{equation} \label{eq:qqq}
		L = \sum_{k\in F}l_{k}^{q},
	\end{equation}
	where $l_k$ denotes the brightness of pixel $k$ in the image.
	
	\item Then, the image is evenly divided into multiple rectangular patches. 	Similar to the calculation of $L$,  for each patch $F_i$, its {\it modified brightness} $L_i$ is given by:
	\begin{equation}
		L_i = \sum_{k\in F_i} l_k^q.
	\end{equation} 
	The number $N_i$ of pixels selected from patch $F_i$ is set to be proportional to the modified brightness $L_i$ of $F_i$:
	\begin{equation}
		N_i = \left( \frac{L_i}{L} \right) N_\sigma.
	\end{equation}
	\item	Finally, the brightest $N_i$ pixels in patch $F_i$ are selected as the representative pixels. .
\end{enumerate}

\paragraph{Applying GW-based Methods}
After the selection of representative pixels, the GW-based methods (e.g., GW~\cite{GW}, SoG~\cite{SoG}, GGW~\cite{GGW}, GE$_1$~\cite{GE} and GE$_2$~\cite{GE}) are applied to generate the final  estimation. When applying GW or SoG, for example, we only need to calculate the Minkowsky-norm of selected pixels to obtain the final estimation. Whereas for GGW, GE$_1$ and GE$_2$, the derivative or Gaussian filter has to be performed on the downsampled image before selecting the pixels of interest, and the Minkowsky-norm of selected pixels is then computed.  An example illustrating the major steps of the PBP algorithm is given in Figure~\ref{illustration}
 
\paragraph{Remarks}	 
	The way we select bright pixels can be viewed as a modification of that in the BP method~\cite{Bright Pixels}. The difference lies in that we select more local bright pixels while abandoning some global bright pixels. In this way, the performance of algorithm can be improved, especially when dealing with images where the brightest areas are colored objects, or those with multi-illuminants. However, the alteration also requires to pay special attention to the spatial dispersion of the selected pixels. Generally speaking,  high dispersion implies that more local bright pixels are chosen, to some extent missing the good properties of global bright pixels in predicting the illuminant. Conversely, low dispersion of the selected pixels can boost the risk of failure, as what happened to  BP~\cite{Bright Pixels}. Therefore, it is of great importance to adjust the dispersion so that the aforementioned negative effects could be minimized.

	Fortunately, the two parameters in PBP, i.e., i) the number of patches and ii) the power $q$ in~\eqref{eq:qqq}, have direct control
on the dispersion of selected pixels. For certain regions, the larger the number of patches is, the more evenly the selected pixels are distributed over the image; The higher the power $q$ is, the more sampling is on the patches with high brightness, which means that the selected pixels are more concentrated in bright areas. It is worth mentioning that when the patch number is one, our method reduces to the conventional BP method~\cite{Bright Pixels}. 

%

	\begin{figure}[t]
		\centering
		\subfigure[NUS $8$-Camera]{  
     \hspace{-3mm}	\includegraphics[width=42mm]{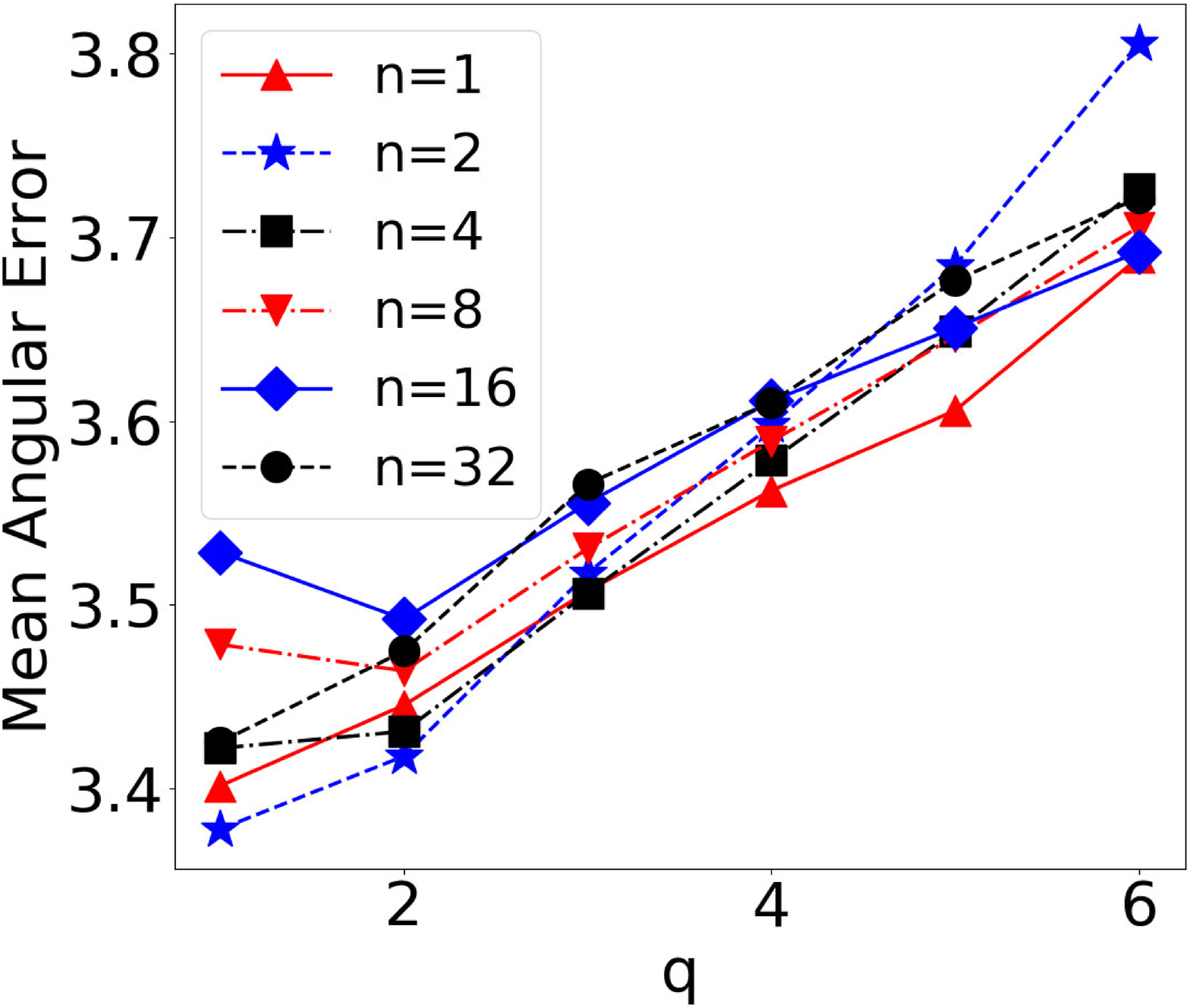}} \hspace{-1mm}
		\subfigure[Gehler-Shi]{  
			\includegraphics[width=42mm]{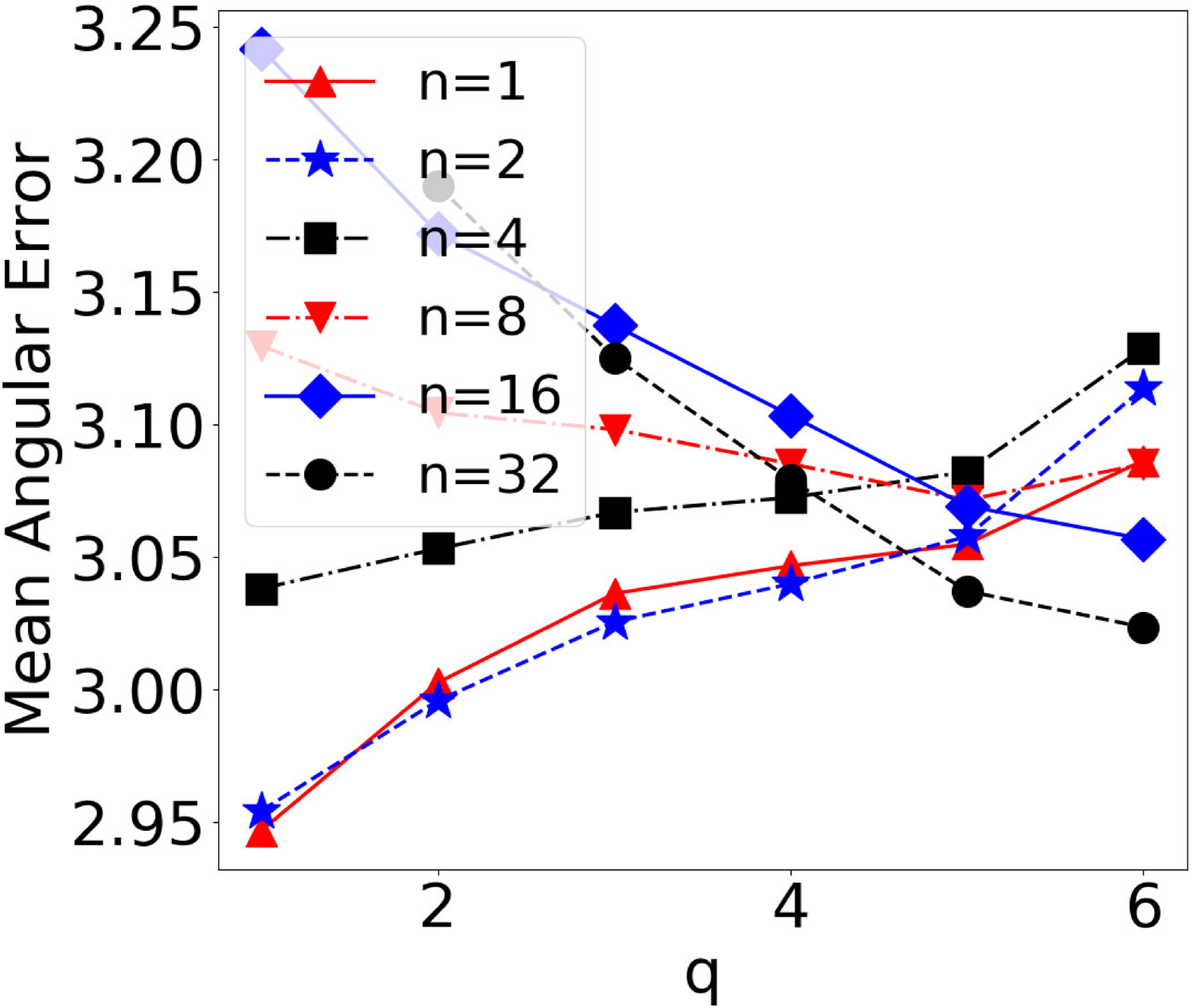} \hspace{-2mm}} 
		\caption{Mean angular error as a function of power $q$ for different $n$ on two datasets.}
		\label{patch and dimension}
	\end{figure}

	\begin{table}[t]
		\centering
		\caption{The parameters $(\sigma, S, p)$ that minimizes the sum of mean and median angular errors. }
		\hspace{-3.6mm}	 \begin{tabular}{c|c|c|c|c|c}\hline
		\hspace{-2mm}		$(n, q)$   \hspace{-2mm}	 & GW & SoG & GGW & GE$_1$ & GE$_2$ 
			\\ \hline  
		\hspace{-2mm}	$(1, 1)$  \hspace{-2mm} &  \hspace{-2mm} (2\%,11,1)  \hspace{-2mm} &  \hspace{-2mm} (0.5\%,4,1)  \hspace{-2mm} &  \hspace{-2mm} (2\%,3,3)  \hspace{-1mm}	 \hspace{-2mm} &  \hspace{-2mm} (4\%,3,1) \hspace{-2mm}	 & \hspace{-2mm}	 (4\%,6,1)  \hspace{-2mm} 
		\\ 
		\hspace{-2mm}	$(2, 1)$  \hspace{-2mm} &  \hspace{-2mm}  \hspace{-2mm} (2\%,9,1)   \hspace{-2mm} &  \hspace{-2mm} (0.5\%,3,1) \hspace{-2mm} &  \hspace{-2mm} (2\%,3,3) \hspace{-2mm} &  \hspace{-2mm} (4\%,3,1) \hspace{-2mm} &  \hspace{-2mm} (4\%,6,1)  \hspace{-2mm}
		\\ \hline
		\end{tabular}
	\label{hyperparameter} 
	\end{table}

		\begin{table}[t]
		\centering
		{\tabcolsep=1pt
		\begin{scriptsize}
		\caption{Performance on the NUS $8$-Camera Dataset~\cite{PCA}. The angular error statistics of learning-based and learning-free methods are collected from GI~\cite{Gray Index} and FFCC~\cite{FFCC}. The PBP algorithm  is denoted as ``PBP - $(n, q)$ + (name of applied GW-Based method)''. The running time of learning-free algorithms is tested on our PC and reported in seconds, averagely per image.  Unreported results are left dash.}
		\begin{tabular}{|l|ccccc|c|c|}\hline
			\multirow{2}*{Algorithm} & \multirow{2}*{Mean} & \multirow{2}*{Med.} & \multirow{2}*{Tri.} & Best & Worst & {Geo.} & \multirow{2}*{Time} \\
			&&&& $25\%$ & $25\%$ & Mean & \\\hline
			\multicolumn{6}{|c|}{\bf{\emph{Learning-based}}} && \\
			Pixel-based Gamut~\cite{Gamut} & 5.27 & 4.26 & 4.45 & 1.28 & 11.16 & 4.27 &\textemdash\\
			Edge-based Gamut~\cite{Gamut} & 4.40 & 3.30 & 3.45 & 0.99 & 9.83 & 3.45 &\textemdash\\
			Natural Image Statistics~\cite{Natural Image Statistics}& 3.45 & 2.88 & 2.95 & 0.83 & 7.18 & 2.80 &\textemdash\\
			Bayesian revisited~\cite{Bayesian revisited} & 3.50 & 2.36 & 2.57 & 0.78 & 8.02 & 2.66 &\textemdash\\
			Spatio-spectral(GenPrior)~\cite{Spatio-spectral} & 3.06 & 2.58 & 2.74 & 0.87 & 6.17 & 2.59 &\textemdash\\
			Corrected-Moment(19 Edge)~\cite{Corrected Moment} & 3.03 & 2.11 & 2.25 & 0.68 & 7.08 & 2.33 &\textemdash\\
			Corrected-Moment(19 Color)~\cite{Corrected Moment}& 3.05 & 1.90 & 2.13 & 0.65 & 7.41 & 2.26 &\textemdash\\
			DS-Net(HypNet+SelNet)~\cite{Shi2016}& 2.24 & 1.46 & 1.68 & 0.48 & 6.08 & 1.74 &\textemdash\\
			CCC(dist+ext)~\cite{CCC} & 2.38 & 1.48 & 1.69 & 0.45 & 5.85 & 1.73 &\textemdash\\
			FC$_4$~\cite{FC4}  & 2.12 & 1.53 & 1.67 & 0.48 & 4.78 & 1.66 &\textemdash\\
			Cheng {\it et~al.} 2015~\cite{Cheng 2015} & 2.18 & 1.48 & 1.64 & 0.46 & 5.03 & 1.65 &\textemdash\\
			FFCC~\cite{FFCC} & {\bf 1.99} & {\bf 1.31} & {\bf 1.43} & {\bf 0.35} & {\bf 4.75} & {\bf 1.44} &\textemdash\\\hline
			
			\multicolumn{6}{|c|}{\bf{\emph{Learning-free}}} && \\
			White Patch~\cite{WP} & 9.91 & 7.44 & 8.78 & 1.44 & 21.27 & 7.24 & 0.057\\
			Gray World~\cite{GW} & 4.59 & 3.46 & 3.81 & 1.16 & 9.85 & 3.70 & 0.082\\
			Shades of Gray~\cite{SoG} & 3.67 & 2.94 & 3.03 & 0.99 & 7.75 & 3.02 & 0.555\\
			LSRS~\cite{LSRS} & 3.45 & 2.51 & 2.70 & 0.98 & 7.32 & 2.79 & 0.199\\
			$2$nd-order Gray Edge~\cite{GE} & 3.36 & 2.70 & 2.80 & 0.89 & 7.14 & 2.76 & 0.961\\
			$1$st-order Gray Edge~\cite{GE} & 3.35 & 2.58 & 2.76 & 0.79 & 7.18 & 2.67 & 0.801\\
			General Gray World~\cite{GGW2} & 3.20 & 2.56 & 2.68 & 0.85 & 6.68 & 2.63& 0.565\\
			Cheng {\it et~al.} 2014~\cite{PCA} & 2.93 & 2.33 & 2.42 & 0.78 & {\bf 6.13} & 2.40 & 0.530\\
			Gray Index~\cite{Gray Index} & {\bf 2.91} & {\bf 1.97} & {\bf 2.13} & {\bf 0.56} & 6.67 & {\bf 2.15} &  0.961\\\hline
			\multicolumn{6}{|c|}{\bf{\emph{Proposed}}} && \\
			PBP - (1, 1) + GW & 2.89 & 2.02 & 2.21 & 0.62 & 6.60 & 2.21 &{\bf  0.0021} \\
			PBP - (1, 1) + SoG & {\bf 2.76} & {\bf 1.99} & {\bf 2.16} & 0.61 & {\bf 6.25} & {\bf 2.14} & 0.014\\
			PBP - (1, 1) + GGW & 2.90 & 2.04 & 2.21 & 0.62 & 6.65 & 2.21 & 0.028\\
			PBP - (1, 1) + GE$_1$ & 3.83 & 3.07 & 3.24 & 1.03 & 7.93 & 3.15& 0.051\\
			PBP - (1, 1) + GE$_2$ & 3.03 & 2.15 & 2.35 & 0.72 & 6.86 & 2.37 & 0.014\\
			PBP - (2, 1) + GW & 2.90 & 2.07 & 2.27 & {\bf 0.59} & 6.64 & 2.21 & 0.0028\\
			PBP - (2, 1) + SoG & 2.81 & 2.03 & 2.22 & 0.63 & 6.31 & 2.18 & 0.020\\
			PBP - (2, 1) + GGW & 2.90 & 2.04 & 2.21 & 0.62 & 6.65 & 2.21 & 0.024\\
			PBP - (2, 1) + GE$_1$ & 3.83 & 3.07 & 3.24 & 1.03 & 7.93 & 3.15 & 0.045\\
			PBP - (2, 1) + GE$_2$ & 3.03 & 2.15 & 2.35 & 0.71 & 6.86 & 2.36 & 0.015\\\hline
		\end{tabular}
		\label{outcome_nus}
		\end{scriptsize}}  
	\end{table}

	\begin{table}[t]
		\centering
		{\tabcolsep=1pt
		\begin{scriptsize}
		\caption{Performance on the Gehler-Shi Dataset~\cite{Gehlershi}.} 
		\begin{tabular}{|l|ccccc|c|c|}\hline
			\multirow{2}*{Algorithm} & \multirow{2}*{Mean} & \multirow{2}*{Med.} & \multirow{2}*{Tri.} & Best & Worst & Geo. & \multirow{2}*{Time} \\
			&&&& 25\% & 25\% & Mean & \\\hline
			\multicolumn{6}{|c|}{\bf{\emph{Learning-based}}} &&\\
			Edge-based Gamut~\cite{Gamut} & 6.52 & 5.04 & 5.43 & 1.90 & 13.58 & 5.40 &\textemdash\\
			Bayesian revisited~\cite{Bayesian revisited} & 4.82 & 3.46 & 3.88 & 1.26 & 10.49 & 3.86 &\textemdash\\
			Natural Image Statistics~\cite{Natural Image Statistics}& 4.19 & 3.13 & 3.45 & 1.00 & 9.22 & 3.34 &\textemdash\\
			Spatio-spectral Statistics~\cite{Spatio-spectral} & 3.59 & 2.96 & 3.10 & 0.95 & 7.61 & 2.99 &\textemdash\\
			Pixel-based Gamut~\cite{Gamut} & 4.20 & 2.33 & 2.91 & 0.50 & 10.72 & 2.73 &\textemdash\\
			Exemplar-based~\cite{Exemplar-based} & 2.89 & 2.27 & 2.42 & 0.82 & 5.97 & 2.39 &\textemdash\\
			Corrected-Moment~\cite{Corrected Moment}& 2.86 & 2.04 & 2.22 & 0.70 & 6.34 & 2.25 &\textemdash\\
			Chakrabarti {\it et~al.} 2015~\cite{Cha} & 2.56 & 1.67 & 1.89 & 0.52 & 6.07 & 1.91 &\textemdash\\
			Cheng {\it et~al.} 2015~\cite{Cheng 2015} & 2.42 & 1.65 & 1.75 & 0.38 & 5.87 & 1.73 &\textemdash\\
			CCC~\cite{CCC} & 1.95 & 1.22 & 1.38 & 0.35 & 4.76 & 1.40 &\textemdash\\
			Shi {\it et~al.} 2016~\cite{Shi2016}& 1.90 & 1.12 & 1.33 & 0.31 & 4.84 & 1.34 &\textemdash\\
			FC$_4$ (SqueezeNet)~\cite{FC4}  & {\bf 1.65} & 1.18 & 1.27 & 0.38 & {\bf 3.78} & 1.29 &\textemdash\\
			FFCC~\cite{FFCC} & 1.78 & {\bf 0.96} & {\bf 1.14} & {\bf 0.29} & 4.62 & {\bf 1.21} &\textemdash\\\hline
			\multicolumn{6}{|c|}{\bf{\emph{Learning-free}}} &&\\
			White Patch~\cite{WP} & 7.55 & 5.68 & 6.35 & 1.45 & 16.12 & 5.76 & 0.040\\
			Gray World~\cite{GW} & 6.36 & 6.28 & 6.28 & 2.33 & 10.58 & 5.73 & 0.060\\
			$1$st-order Gray Edge~\cite{GE} & 5.33 & 4.52 & 4.73 & 1.86 & 10.03 & 4.63 & 0.564\\
			$2$nd-order Gray Edge~\cite{GE} & 5.13 & 4.44 & 4.62 & 2.11 & 9.26 & 4.60 & 0.683\\
			Shades of Gray~\cite{SoG} & 4.93 & 4.01 & 4.23 & 1.14 & 10.20 & 3.96 & 0.411\\
			General Gray-World~\cite{GGW2} & 4.66 & 3.48 & 3.81 & 1.00 & 10.09 & 3.62 & 0.419\\ 
			LSRS~\cite{LSRS} & 3.31 & 2.80 & 2.87 & 1.14 & {\bf 6.39} & 2.87 & 0.141\\
			Cheng {\it et~al.} 2014~\cite{PCA} & 3.52 & 2.14 & 2.47 & 0.50 & 8.74 & 2.41 & 0.354\\
			Gray Index~\cite{Gray Index} & {\bf 3.07} & {\bf 1.87} & {\bf 2.16} & {\bf 0.43} & 7.62 & {\bf 2.10} & 0.673\\\hline
						\multicolumn{6}{|c|}{\bf{\emph{Proposed}}} &&\\
			PBP - (1, 1) + GW & 3.40 & 2.11 & 2.43 & 0.46 & 8.54 & 2.32 & {\bf 0.0018} \\\hline
		\end{tabular}
		\label{outcome_gehlershi}
		\end{scriptsize}}  
	\end{table}

	\begin{figure*}[t]
		\centering
		\includegraphics[width=\linewidth]{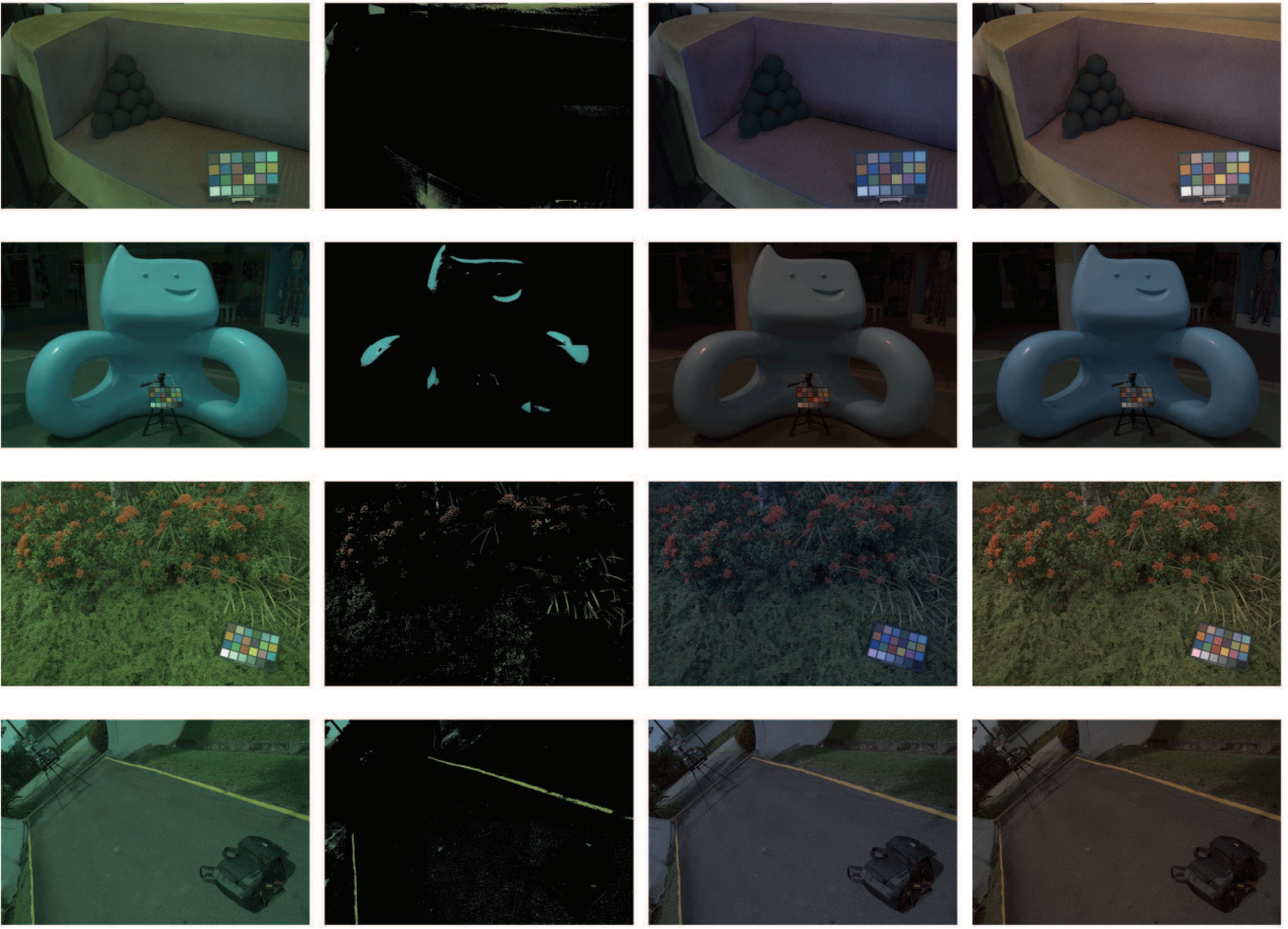}
		\caption{Failure cases of PBP. From left to right: the input image, selected pixels (sampling rate = $3\%$), image corrected with the predicted illuminant, and image corrected with the groundtruth illuminant. A gamma correction with $\gamma=2.2$ is used to enhance the overall brightness of images.}
		\label{Failure Case}
	\end{figure*}

%
	
	\section{Evaluation} \label{sec:exp}
\paragraph{Setup} We empirically evaluate the performance of our algorithm on the NUS $8$-Camera Dataset~\cite{PCA} and the Gehler-Shi Dataset~\cite{Gehlershi}. Note that both the Gehler-Shi Dataset~\cite{Gehlershi} and NUS $8$-Camera Dataset~\cite{PCA} are mainly composed of images with aspect ratio $3:2$ (some images in the NUS $8$-Camera Dataset are with ratio $4:3$). Thus, we cut the images into $3n \times 2n$ patches for some positive integer $n$, so that the shape of each patch is approximately square.  
	
	In order to  find out a satisfactory pair of parameters $(n, q)$, we perform a grid search over $ n\in\{1, 2,4,8,16,32\}~\text{and}~q \in\{1,2,3,4,5,6\}.$ As shown in Figure~\ref{patch and dimension}, the results of these two datasets reveal different patterns. Specifically, for the NUS $8$-Camera Dataset, the mean angular error increases with  $q$ for $n\leq 8$ but decreases with  $q$ for $n\geq 16$. Whereas for the Gehler-Shi Dataset, the mean angular error generally increases with  $q$ for all region of $n$. Despite the disparities, it can be observed that PBP achieves the best performance when $(n, q) \in \{ (1,1), (2,1) \}.$ Therefore, we use these two sets of parameters for the PBP algorithm in our experiment. 
	
Based on the NUS $8$-Camera Dataset, we further conduct a grid search for a proper set of parameters $(\sigma, S, p)$, where $\sigma$ is the overall sampling rate of bright pixels (see equation~\eqref{eq:sigma} for details), $S$  the downsampling interval, and $p$ is the order of Minkowsky-norm used in the GW-based methods (e.g.,~\cite{GW, WP, SoG, GGW, GGW2, GE}; See Section~\ref{sec:algbpb}). 
Specifically, for each GW-based algorithm (e.g., GW, SoG, GGW, GE$_1$ or GE$_2$) and each pair of parameters $(n, q) \in \{(1,1), (1,2)\}$, we individually search for a parameter set that minimizes the sum of mean and median angular error. The range of grid search is set to be $\sigma \in \{2\%, 3\%, 4\%\},~S \in \{3,4,5,6,7,8,9,10,11,12\}$ and $p \in \{1,2,3\}$. However, for SoG, we adjust $\sigma$ in a wider range (from $0.25\%$ to $5\%$) in order to further explore the influence of the sampling rate to our algorithm. The results of our grid search are given in Table~\ref{hyperparameter}.

\paragraph{Results}	With the determined values of parameters, we evaluate the angular error performance and running time of PBP. We also compare the results with those of other illuminant estimation methods in terms of some popular metrics, i.e., the mean, median, trimean, mean of the best $25\%$, mean of the worst $25\%$ angular error, and their geometric mean. The results on the NUS $8$-Camera Dataset~\cite{PCA} and Gehler-Shi Dataset~\cite{Gehlershi} are shown in Tables \ref{outcome_nus} and~\ref{outcome_gehlershi}, respectively. Overall, it can be observed that for both datasets, FFCC~\cite{FFCC} performs the best among  all approaches under test.  	
In particular, PBP outperforms all the learning-free methods except for GI~\cite{Gray Index}, which performs worse than ours for the  NUS $8$-Camera Dataset and slightly better on the Gehler-Shi Dataset. Even when compared to the learning-based methods, PBP still exhibits competitive performance. Indeed, it has superior performance to a series of popular learning-based methods, such as the Spatio-spectral Statistics~\cite{Spatio-spectral} and Corrected Moment~\cite{Corrected Moment}. 
	
	
	The running time for the learning-free methods  is also given in Tables~\ref{outcome_nus} and~\ref{outcome_gehlershi}, where the parameters of those mathods are specified in Table~\ref{downsample_parameter}. It is based on an Intel Core i$7$-$6820$HQ using the Python $3.6$ implementation. One  can observe that the running time of GW-based methods increases sharply with the order of Minkowsky-norm  $p$. Remarkably, the PBP algorithm (with various GW-based method) takes uniformly less than $50$ms to process a $1080$p image. The fastest speed can even reach $2$ms/image, which is roughly $40$ times faster than the conventional GW method and hundreds of times faster than Cheng {\it et~al.}~\cite{PCA} and GI~\cite{Gray Index}. Table~\ref{hyperparameter},~\ref{outcome_nus} and~\ref{outcome_gehlershi} also indicate that the larger the downsampling interval is, the faster the PBP algorithm can be. 

	\paragraph{Failure Case Study} The failure cases of PBP, in which the angular errors are above $10^{\circ}$, are displayed in Figure~\ref{Failure Case}. As shown in this figure, the stereotypical images, for which our method fails to predict the illuminant, have no obvious highlights, light sources or white surfaces. For those cases, the PBP algorithm may select pixels that reflect relatively bright non-white objects. The third row of Figure~\ref{Failure Case} is one such example. Specifically, the pixels of bright-color flowers and those of green plants are selected, thus inducing bias in the estimation of light source of the scene.
	
	\section{Conclusion}
	In this paper, we have developed an efficient color constancy algorithm called PBP, which is based on both the bright pixels and patch brightness. Evaluation on various benchmark datasets have demonstrated that the PBP algorithm has very competitive performance compared to the state-of-the-art learning-free approaches, yet can be hundreds of times faster. 
	 Therefore, our algorithm is particularly useful for the practical applications where the computational resource is limited.
	
	We would like to mention that our algorithm may be useful for the design of full-screen smart phone. Nowadays, there have been various solutions for the full-screen smart phone, such as the notch screen and the water drop screen. For those screens, however, their screen-to-body ratio does not really reach $100\%$. To realize the $100\%$ screen-to-body ratio, we can place the front camera of smart phone behind a transparent or semi-transparent screen. When the screen is lighted, the video taken by this camera will naturally be colored by the illuminant caused from what is displayed on the screen. Perfectly removing the color bias from the video via PBP and thereby realizing a real full-screen display of smart phone can be an interesting and challenging job. Our future work will be directed towards this avenue. 
	
{\small
	\bibliographystyle{ieee_fullname}
	
}

\end{document}